\documentclass[10pt,conference]{IEEEtran}
\IEEEoverridecommandlockouts

\usepackage{xspace}
\usepackage{tcolorbox}
\usepackage{enumitem}
\usepackage[ruled]{algorithm2e}
\usepackage{xcolor}
\definecolor{wine}{RGB}{181,29,25}
\usepackage[numbers,sort]{natbib}
\usepackage[colorlinks=true,citecolor=wine,linkcolor=wine,urlcolor=wine,bookmarks=false]{hyperref}

\usepackage{booktabs} 
\usepackage{multirow}
\usepackage{graphicx}
\usepackage{tabularx}
\newcolumntype{C}[1]{>{\centering\arraybackslash}p{#1}}
\usepackage{colortbl}
\usepackage{graphicx}  
\usepackage{float}  
\usepackage{subfig}
\usepackage{makecell} 
\usepackage{color}
\usepackage{arydshln} 
\usepackage{caption}
\usepackage{wrapfig}
\usepackage{colortbl}
\usepackage{amsthm} 
\usepackage{amsmath}
\usepackage{adjustbox}
\usepackage{threeparttable}
\usepackage{stfloats}
\usepackage{pifont}
\usepackage{xurl}
\usepackage{amsfonts}
\usepackage{circledsteps}

\theoremstyle{definition}

\usepackage{microtype}
\newcommand{\ie}[0]{\textit{i.e.,}\xspace}

\newcommand{\tool}{\textsc{SkillSentry}\xspace}

\newcommand{\todo}[1]{\textcolor{black}{#1}}
\newcommand{\revision}[1]{\textcolor{black}{#1}}
\newcommand{\ly}[1]{\textcolor{black}{#1}}

\def\BibTeX{{\rm B\kern-.05em{\sc i\kern-.025em b}\kern-.08em
    T\kern-.1667em\lower.7ex\hbox{E}\kern-.125emX}}
\begin{document}

\title{\tool: Reliable Skill Execution for LLM Agents via Runtime Assurance}

\author{\IEEEauthorblockN{
You Lu\IEEEauthorrefmark{1},
Xinyu Huang\IEEEauthorrefmark{1}, 
Bihuan Chen\IEEEauthorrefmark{1},
Xin Peng\IEEEauthorrefmark{1}}
\IEEEauthorblockA{\IEEEauthorrefmark{1}College of Computer Science and Artificial Intelligence, China}}

\maketitle

\begin{abstract}
LLM agents are increasingly equipped with skills~to perform complex tasks through multi-step reasoning and tool~use. Although skills provide reusable procedural knowledge, agents may still execute them unreliably. Even when an agent has demonstrated the capability to complete tasks under the guidance of a skill, it may fail to do so consistently across similar tasks or repeated runs due to \ly{deviations from the skill procedure or incorrect execution of individual steps.} Such instability limits the practical reliability of LLM agents. \ly{To address this~problem, we propose \tool, a skill-oriented runtime assurance framework built upon a new domain-specific language (DSL) for representing runtime guidance for skill execution. \tool initializes the runtime guidance by combining a skill specification extracted from the corresponding skill document with execution experience mined from historical successful and failed traces. It then wraps around the agent execution loop to monitor and guide skill execution under the current guidance, while iteratively refining the guidance using newly collected traces.} We evaluate \tool on 15 skills across two LLM agents, each paired with two backbone models, \ie Claude Code with Claude-Haiku-4.5 and Claude-Opus-4.6, and Codex with GPT-5.2 and GPT-5.4. \ly{Our results show that \tool improves the task success rate of LLM agents by \todo{24.1\%} across skills, on average, while exhibiting lower variability across repeated runs.}

\end{abstract}

\section{Introduction}\label{sec:intro}

Large language models (LLMs)~\cite{chatgpt,gemini,glm5,claudehaiku} have shown strong capabilities in understanding user instructions, generating code, and solving complex reasoning questions. Building on these capabilities, LLM agents~\cite{yao2022react,schick2023toolformer,qin2024toolllm,codex,claudecode} further extend LLMs from passive text generation to autonomous task execution, \ie planning intermediate steps, invoking external tools, observing execution feedback, and refining their actions~\cite{wu2024autogen,yang2024sweagent,bouzenia2025repairagent,zhang2024autocoderover, hou2024large}. As user tasks become more complex and diverse, skills~\cite{li2026skillsbench,zhou2026skillgenbench,zhong2026skilllearnbench} have emerged as an important abstraction for packaging reusable task-solving knowledge. A skill usually describes how an agent should complete a class of tasks through predefined steps, including high-level procedures, tool-use patterns, and task-specific constraints~\cite{wang2023voyager}. By providing such procedural guidance, skills help LLM agents reuse validated task-solving strategies, constrain the execution space, and improve their capability to complete complex tasks.

However, equipping an LLM agent with a skill does not guarantee reliable execution. Even when the LLM agent has demonstrated the capability to complete a task under the guidance of a skill before, it may fail to reproduce the correct execution consistently in later runs~\cite{gupta2026reliabilitybench,xue2025characterization}. Such failures do not necessarily indicate a lack of fundamental model capability for task-solving. Rather, they often arise from the generative nature of LLM agent execution, where each action is produced from an evolving context of prior reasoning, tool feedback, and intermediate observations. \ly{Thus, small variations during agent execution may lead to deviations from the skill procedure or incorrect execution of individual steps}, resulting in task~failures.

\ly{Existing approaches improve the reliability of LLM agent execution from different perspectives.} Some studies focus on acquiring, generating, or selecting skills for better reusability~\cite{li2026skillsbench,zhou2026skillgenbench}. Some automatic prompt optimization and reflection-based approaches refine prompts, instructions, \ly{in-context memories}, or skills based on execution feedback~\cite{shinn2023reflexion,zhao2023expel,zhou2023large,pryzant2023automatic,yang2024large,ni2026trace2skill}. A few studies improve the underlying model capability through supervised fine-tuning, reinforcement learning, or post-training on agent execution traces~\cite{ouyang2022training,bai2022constitutional,schick2023toolformer,qin2024toolllm,bouzenia2025understanding}. These approaches mainly improve LLM agents by updating \ly{external artifacts or the underlying model parameters}, thereby enhancing the task-solving capability before agent deployment. In contrast, some studies focus on runtime guardrails and enforcement systems~\cite{wang2026agentspec,chen2025shieldagent,kumar2026infrastructuresentinel}, constraining agent behaviors during execution according to user-specified rules, safety policies, or domain constraints. They are effective for preventing unsafe or policy-violating actions, but a skill-based agent may still fail without violating explicit safety policies. \ly{Although prior approaches can improve LLM agent capabilities or constrain unsafe runtime behavior, it remains unclear how to improve the runtime reliability of skill execution when an agent already has the basic capability to complete the task but may still deviate from the skill procedure or execute individual steps incorrectly.}

To bridge this gap, we propose \tool, a skill-oriented runtime assurance framework for improving the~reliability of LLM agent execution. \tool does not aim to teach an LLM agent fundamentally new skills or compensate for tasks beyond the capability of the underlying model. Instead, it targets cases where an agent has demonstrated the capability to complete tasks under the guidance of a skill but may still execute the skill unreliably across similar tasks or repeated~runs.

\ly{\tool is built upon a new domain-specific language (DSL) for representing runtime guidance for skill execution, and consists of an initialization stage and a self-evolving stage. In the initialization stage, it extracts a \textit{skill specification} from each skill document, capturing the expected steps, step dependencies, constraints, and completion requirements. It also mines \textit{execution experience} from historical successful~and failed traces, including \textit{validated action patterns}, \textit{failure-associated action patterns}, and step-level \textit{suggestions} and \textit{warnings}. By combining the skill specification with the mined experience, \tool constructs the initial runtime guidance.}

In the self-evolving stage, the agent executes new task queries under current runtime guidance through the step-aware runtime assurance mechanism. \tool monitors agent execution, delivers step-level suggestions and warnings to improve the execution of individual skill steps, intervenes upon detecting procedural deviations or failure-associated action patterns, and \revision{checks the completion of required steps} before accepting the final output. \revision{The resulting traces are used to refine the execution experience, enabling \tool to iteratively improve the runtime guidance as new execution traces are~collected.}

We evaluate \tool on 15 skills across \ly{two widely used LLM agents, each paired with two backbone models, \ie Claude Code~\cite{claudecode} with Claude-Haiku-4.5~\cite{claudehaiku45} and Claude-Opus-4.6, and Codex~\cite{codex} with GPT-5.2~\cite{gpt52} and GPT-5.4~\cite{gpt54}.} \ly{Our evaluation shows that \tool improves the average task success rate by \todo{24.1\%} across skills, reduces the standard deviation (SD) of task success rates across repeated runs by~\todo{41.1\%}}, and introduces little runtime overhead. \ly{Further self-evolving analysis shows that \tool can iteratively refine the runtime guidance for reliable skill execution.} \ly{Our ablation study confirms the contribution of failure-associated action patterns, suggestions, and warnings in the runtime guidance to the final effectiveness. Step-level guidance delivery is more effective than placing all guidance in the system prompt, increasing the task success rate by \todo{5.8\%} and reducing the SD by \todo{26.8\%}.} Finally, our generalization study shows that the constructed runtime guidance can transfer across different backbone models within the same agent to some~extent. 

The main contributions of this work are as follows:
\begin{figure}[!t]
    \centering
    \includegraphics[width=\linewidth]{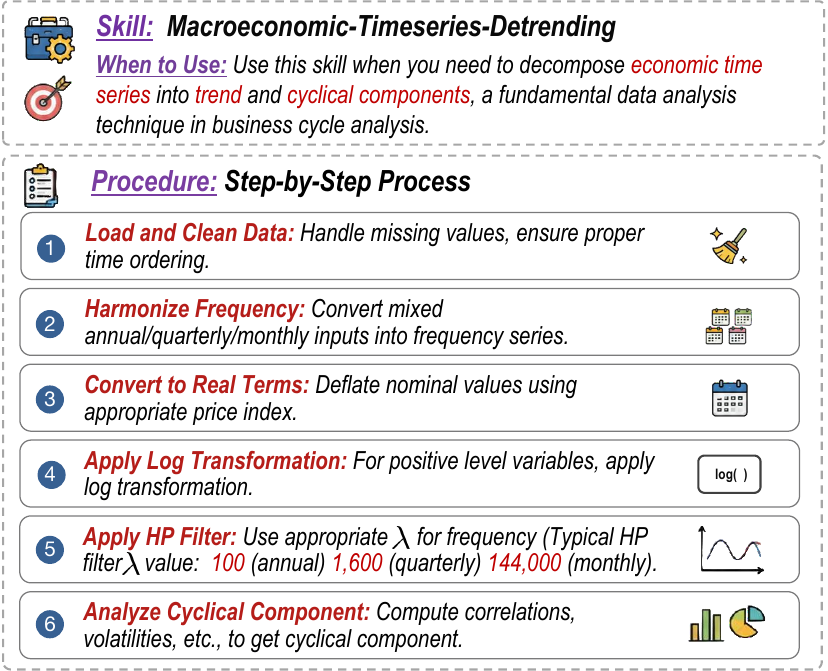}
    \caption{Example Skill for Economic Time Series Detrending}
    \label{fig:skill_example}
    \vspace{-0.5pt}
\end{figure}

\begin{itemize}[leftmargin=*]
    \item \ly{We design a domain-specific language for representing skill-oriented runtime guidance that combines a skill specification extracted from the skill document with execution experience mined from successful and failed agent traces.}
    \item \ly{We propose \tool, a skill-oriented runtime assurance framework that wraps around the agent execution loop, monitors and guides skill execution, and iteratively refines its runtime guidance using newly collected execution traces.}
    \item We implement a prototype of \tool, and conduct experiments to demonstrate its effectiveness and efficiency.
\end{itemize}

\section{Background and Motivation}\label{sec:background-and-motivation}

We first introduce the background of skill-based agent execution, and then introduce the motivation for our work.

\begin{figure}[!t]
    \centering
    \includegraphics[width=\linewidth]{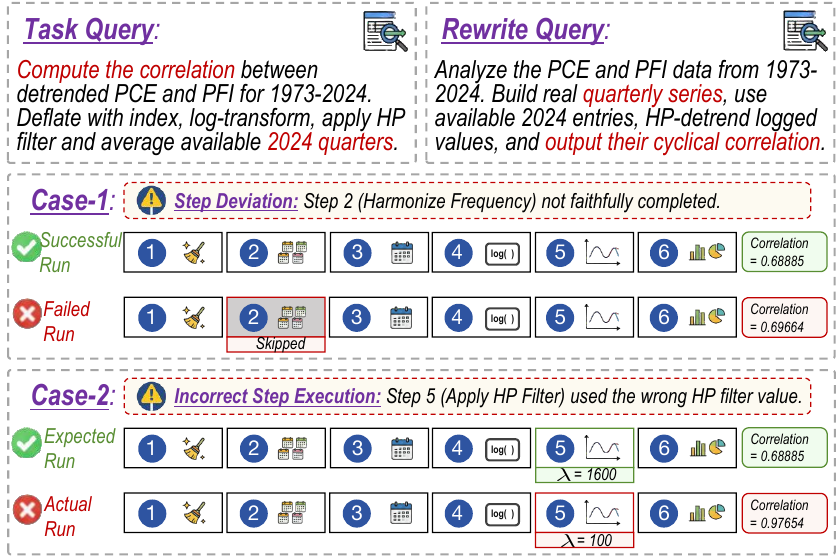}
    \caption{Example Cases of Skill Execution Failure}
    \label{fig:case_example}
\end{figure}

\subsection{Background}
\textbf{Skill-Based Agent Execution.} An LLM agent completes tasks through iterative interactions with external environments~\cite{yao2022react,wu2024autogen,bouzenia2025understanding}. To guide multi-step planning and execution, skills have been introduced as reusable procedural knowledge for completing a class of tasks~\cite{li2026skillsbench,zhong2026skilllearnbench,zhou2026skillgenbench}. A skill~usually contains a name, a description specifying when it should~be used, and a procedure describing the expected execution~steps. Fig.~\ref{fig:skill_example} shows an example skill for macroeconomic timeseries detrending. An LLM agent executes a skill through the agent runtime. The runtime maintains the interaction context, presents the skill and available tools to the model, receives model-generated actions, executes tool calls, and returns observations to the model for the next decision. \ly{Therefore, the abstract procedure described in a skill is realized as a concrete runtime trace comprising multiple iterations of action planning, tool invocation, and execution feedback, followed by a final output.}

\textbf{Hook.} Modern agent systems commonly expose hooks as extension points in the runtime loop~\cite{claudecode,codex}. Hooks can be triggered at different lifecycle points, such as before an action is executed, after execution feedback is returned, or before the agent terminates~\cite{claude_code_hooks,codex_hooks}. \ly{These hooks allow an external module to observe the next action and provide additional information to the agent during execution. In this work, we use hooks to monitor and guide LLM agents' skill execution at runtime.}

\subsection{Motivation}

\ly{Although skills provide procedural knowledge for task completion, runtime decisions depend on iterative reasoning and execution feedback, making the execution process sensitive to small variations in intermediate contexts. Thus, an agent may understand the overall skill but still deviate from the skill procedure or execute individual steps incorrectly.}

To illustrate this problem, we show two cases of Claude Code~\cite{claudecode} under the guidance of \ly{the skill for macroeconomic timeseries detrending. As illustrated by \textit{Case-1} in Fig.~\ref{fig:case_example}}, the task query asks the agent to compute the correlation between the cyclical components of personal consumption expenditure (PCE) and private fixed investment (PFI) from 1973 to 2024. The skill procedure requires the agent to explicitly harmonize mixed-frequency inputs before real-term conversion, log transformation, HP filtering, and correlation analysis. The agent had previously completed the task query successfully, demonstrating its capability to complete such tasks with the skill. \ly{However, later repeated executions of the same task query failed occasionally because it skipped the step of frequency harmonization, deviating from the skill procedure.} 

Besides, we semantically rewrote the query without changing the intended task or expected output. After this benign rewrite, Codex failed to complete the task successfully because it used an incorrect HP filtering parameter. \ly{As illustrated by \textit{Case-2} in Fig.~\ref{fig:case_example}, the skill explicitly specifies $\lambda=1600$ for quarterly data, but the agent applied $\lambda=100$, which~is~intended for annual data. This parameter mismatch changed the extracted cyclical components and eventually led to an~incorrect correlation result.}

\begin{figure*}[!t]
    \centering
    \includegraphics[width=\linewidth]{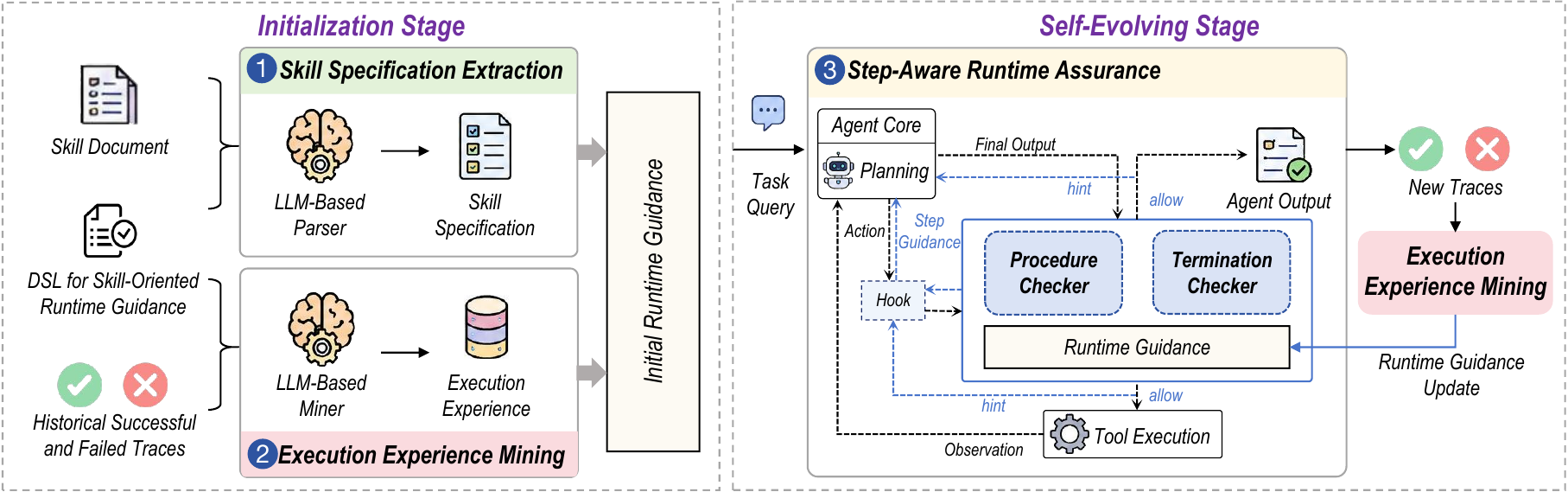}
    \caption{Approach Overview of \tool}
    \vspace{-5pt}
    \label{img:overview}
\end{figure*}

\ly{These cases show that skill-related tasks may still fail even when the agent has previously completed similar tasks successfully under the guidance of corresponding skills.} These observations motivate \tool, which wraps around the LLM agent execution loop, monitors the agent execution progress against the skill procedure, and intervenes when necessary to guide the agent towards reliable skill execution.


\section{Methodology}\label{sec:approach}

We propose \tool, a skill-oriented framework to improve the reliability of LLM agent execution. The approach overview of \tool is presented in Fig.~\ref{img:overview}. \ly{The key idea is to represent the expected skill procedure and historical execution experience as structured runtime guidance, use this runtime guidance to monitor and guide agent skill execution, and iteratively refine it using newly collected execution traces. To this end, we design a domain-specific language (DSL) for runtime guidance that combines a \textit{skill specification} with \textit{execution experience} for each skill (see Sec. \ref{sec:dsl}).}

\ly{Given the DSL, \tool consists of an initialization stage and a self-evolving stage.} In the initialization stage, it first extracts a skill specification (\ie step \Circled{1} in Fig.~\ref{img:overview}) from the corresponding skill document using an LLM-based parser (see Sec.~\ref{sec:skill-specification-extraction}). \ly{The extracted specification captures the expected skill steps, their dependencies, step-level constraints, and the steps required for task completion. \tool then uses historical successful and failed execution traces to mine initial execution experience (\ie step~\Circled{2} in Fig.~\ref{img:overview}). Successful traces provide validated action patterns and step-level suggestions, whereas failed traces reveal failure-associated action patterns and corresponding warnings (see Sec.~\ref{sec:execution-experience-mining}). By combining the extracted specification with the mined experience, \tool constructs the initial runtime guidance for the skill.}

\ly{In the self-evolving stage, the agent executes new task queries under the current runtime guidance through the step-aware runtime assurance mechanism (\ie step~\Circled{3} in Fig.~\ref{img:overview}). \tool wraps around the agent execution loop through runtime hooks, and inspects the actions planned by the agent through a \textit{procedure checker} and a \textit{termination checker}. The procedure checker monitors whether the execution follows the steps embedded in the runtime guidance. When the execution reaches a specific step, it delivers the corresponding step-level suggestions and warnings to help the agent execute the step correctly, and hints the agent to re-plan when procedural deviations or failure-associated action patterns are detected. The termination checker verifies that all required skill steps have been completed before accepting the final output, and prompts the agent to continue planning otherwise (see Sec.~\ref{sec:runtime-assurance}). The resulting new traces are collected for later execution experience mining to refine the runtime guidance. The updated guidance is used in the next round of runtime assurance, forming an iterative optimization loop that progressively improves the runtime guidance for reliable skill execution.}

\subsection{DSL for Skill-Oriented Runtime Guidance}\label{sec:dsl}
\ly{Although skill documents provide procedural knowledge described in natural language, they do not explicitly expose the information needed for runtime procedure checking. Moreover, agent execution traces are typically unstructured. The key challenge is to transform these unstructured data into structured representations that can be consumed by our runtime assurance mechanism.} To this end, \tool introduces a domain-specific language (DSL) for skill-oriented runtime~guidance.

\begin{figure}[!t]
    \centering
    \small
    \resizebox{\columnwidth}{!}{$
    \begin{aligned}
    \langle Guidance \rangle ::= {}&
        \ \textbf{skill}\ \langle SkillName \rangle;\
        \textbf{steps}\ \langle Step \rangle^{+};\ 
        \\
        &\ \textbf{termination}\ \langle StepId \rangle^{+}
        \\
    \langle Step \rangle ::= {}&
        \ \textbf{step\_id}\ \langle StepId \rangle;\
        \textbf{description}\ \langle Text \rangle;\
        \\
        &\ \textbf{depends\_on}\ \langle StepId \rangle^{*};\
        \\
        &\ \textbf{constraints}\ \langle Constraint \rangle^{*};\
        \\
        &\ \textbf{logical\_actions}\ \langle LogicalAction \rangle^{*};
        \\
        &\ \textbf{on\_enter}\ \langle OnEnter \rangle;\
        \\
        &\ \textbf{failure\_patterns}\ \langle FailurePattern \rangle^{*}
        \\
    \langle Constraint \rangle ::= {}&
        \ \langle ToolConstraint \rangle
        \mid \langle ParameterConstraint \rangle
        \\
    \langle ToolConstraint \rangle ::= {}&
        \ \textbf{tool}\ \langle ToolName \rangle
        \\
    \langle ParameterConstraint \rangle ::= {}&
        \ \textbf{parameter}\ \langle ParameterName \rangle;\
        \\
        &\ \textbf{requirement}\ \langle Text \rangle
        \\
    \langle LogicalAction \rangle ::= {}&
        \ \textbf{action\_id}\ \langle ActionId \rangle;\
        \\
        &\ \textbf{patterns}\ \langle ActionPattern \rangle^{+}
        \\
    \langle ActionPattern \rangle ::= {}&
        \ \textbf{tool}\ \langle ToolName \rangle\
        [\textbf{command\_match}\ \langle Regex \rangle]
        \\
    \langle OnEnter \rangle ::= {}&
        \ \textbf{suggestions}\ \langle Text \rangle^{*};\
        \textbf{warnings}\ \langle Text \rangle^{*}
        \\
    \langle FailurePattern \rangle ::= {}&
        \ \langle ActionPattern \rangle;\
        \textbf{reason}\ \langle Text \rangle
    \end{aligned}
    $}
    \caption{Abstract Syntax of Runtime Guidance.}
    \label{fig:dsl-syntax}
    \vspace{-5pt}
\end{figure}

\begin{figure}[!t]
    \centering
    \includegraphics[width=0.9\linewidth]{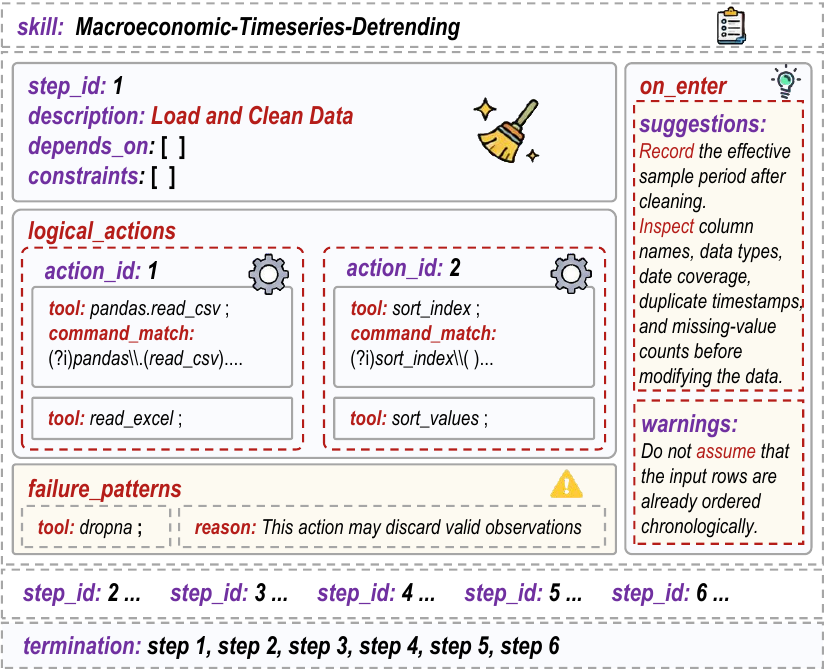}
    \caption{An Example of Runtime Guidance}
    \label{fig:dsl-example}
    \vspace{-0.5pt}
\end{figure}

The DSL combines the skill specification extracted from the skill document and execution experience mined from successful and failed agent execution traces. Fig.~\ref{fig:dsl-syntax} shows the abstract syntax of our DSL. Here, $X^{*}$ denotes zero or more occurrences of $X$, $X^{+}$ denotes one or more occurrences of $X$, and $[\cdot]$ denotes an optional~field. \ly{Fig.~\ref{fig:dsl-example} shows an example of runtime guidance for \revision{the \textit{macroeconomic-timeseries-detrending} skill.}}

\ly{A runtime guidance instance is associated with one skill and consists of a set of procedure steps and termination requirements. Each step is marked by a unique identifier and a short \emph{description}. The \emph{depends\_on} field records prerequisite steps, determining the dependencies between steps, while the \emph{constraints} field records explicit requirements on step execution. We consider two types of constraints, \ie a tool constraint specifies a tool required by the step, and a parameter constraint specifies a required parameter value or condition. The \emph{termination} field lists the required steps that must be completed before the final output can be accepted. All this information can be extracted from the skill document (see~Sec.~\ref{sec:skill-specification-extraction}).}

\ly{Each step further contains a set of \emph{logical\_actions}. A step may involve multiple logical actions, and each logical action may have multiple alternative action patterns. Each pattern specifies an observable tool call and an optional regular expression in the \emph{command\_match} field that matches commands realizing the action. Matching any action pattern indicates that the corresponding logical action has been performed, whereas a step is considered completed only after all of its logical actions have been matched during runtime execution. The \emph{on\_enter} field stores suggestions and warnings for step execution, which are delivered to the agent to support more reliable step execution when the execution reaches the corresponding step. Each step also contains a set of \emph{failure\_patterns}, which~record failure-associated action patterns together with reasons explaining their potential risks. All this information can be mined from successful and failed agent execution traces~(see Sec.~\ref{sec:execution-experience-mining}).}

\subsection{Skill Specification Extraction}\label{sec:skill-specification-extraction}
Given the DSL, \tool first extracts the skill specification from the corresponding skill document. A skill document is written in natural language. Therefore, \tool uses an LLM-based parser with GPT-5.4~\cite{gpt54} to transform each skill document into the specification fields defined by the DSL.

Specifically, given a skill document $\mathcal{D}_{skill}$, the parser extracts four types of specification information. First, it identifies the expected procedure \emph{steps}, assigning each step a unique identifier and a short natural-language description. Second, it identifies prerequisite relations among the steps and stores them in the \emph{depends\_on} fields. Third, it extracts possible tool and parameter constraints for individual steps. Finally, the parser extracts steps that must be completed before the final output can be accepted and stores them as required steps in the \emph{termination}~field.

The parser is provided with the DSL syntax and instructed to preserve the semantics of $\mathcal{D}_{skill}$ without introducing requirements not stated in the original document. All parser prompts are available at our website~\cite{website}. We further perform lightweight structural validation on the parsed specification, including checking the uniqueness of step identifiers, the validity of referenced dependencies, the acyclicity of the dependency graph, the validity of constraint types, and the inclusion of all required steps in the extracted step set. If any check fails, \tool returns the validation errors to the parser, and asks it to revise the result. The resulting skill specification forms the procedural backbone of runtime guidance, while other fields in the DSL are subsequently populated and refined from successful and failed agent execution traces.

\subsection{Execution Experience Mining}\label{sec:execution-experience-mining}

\ly{\tool mines execution experience from successful and failed agent execution traces to populate or refine the experience fields in the runtime guidance.} Let $\mathcal{G}$ denote~the current runtime guidance for a skill, and let $\mathcal{T}_{+}$ and $\mathcal{T}_{-}$ denote the successful and failed traces of agent execution under the skill, respectively. Although $\mathcal{G}$ specifies the expected skill procedure, it may not provide sufficient observable evidence for tracking individual actions or step-level experience for guiding execution of individual steps. Therefore, \tool uses an LLM-based miner powered by GPT-5.4 to analyze $\mathcal{T}_{+}$ and $\mathcal{T}_{-}$ together with $\mathcal{G}$ and the DSL, populating or refining the \emph{logical\_actions}, \emph{on\_enter}, and \emph{failure\_patterns} fields for each step in $\mathcal{G}$. The resulting~runtime~guidance~is~denoted~by~$\mathcal{G}^{*}$.

The \ly{mining} process consists of five stages: \textit{action pattern extraction}, \textit{trace summarization}, \textit{experience field diagnosis}, \textit{experience field edition}, and \textit{guidance validation}. All prompts used by the miner are available at our website~\cite{website}.

\textbf{\revision{Action} Pattern Extraction.} 
\ly{The miner first aligns observable actions in each trace with the skill steps defined in $\mathcal{G}$. The alignment considers the purpose, inputs, outputs, and execution order of each action in the trace, while actions unrelated~to~any skill step remain unmatched.} For each successful trace $\tau_{+}\in\mathcal{T}_{+}$, \ly{the miner extracts validated action patterns associated with each aligned skill step, \ie tool calls of each step. We refer to them as \emph{validated action patterns} because they were exercised in successful executions and therefore provide empirical evidence that the corresponding actions can contribute to completing the step and the overall task. The miner then organizes these patterns into candidate \emph{logical\_actions}, \revision{denoted as $\mathcal{P}_{+}$}, of~the corresponding step, and generates regular expressions if needed for command matching at runtime. Since one step may require multiple logical actions and one logical action may admit multiple valid implementations, patterns that realize the same logical action are grouped as alternatives, while different logical actions remain separate requirements of the step.}

\ly{For each failed trace $\tau_{-}\in\mathcal{T}_{-}$, the miner identifies~observable tool calls associated with the failure in the context of an aligned step. These actions are organized into candidate \emph{failure\_patterns}, \revision{denoted as $\mathcal{P}_{-}$}, of the corresponding step, along with the reasons explaining their potential~risks.}

\textbf{Trace Summarization.} 
Raw execution traces contain long reasoning, tool outputs, file operations, and repeated trials. To reduce noise, we summarize each trace with respect to the steps in $\mathcal{G}$. For each trace $\tau$, the miner produces~a~summary $\Sigma(\tau, \mathcal{G}) = \langle S_{\mathrm{align}}, S_{\mathrm{skip}}, A_{\mathrm{unalign}}, s^{\mathrm{fail}}, y \rangle$, where $S_{\mathrm{align}}$ denotes the steps \revision{aligned} in the trace, $S_{\mathrm{skip}}$ denotes the steps that should have been executed but are not supported by the trace, $A_{\mathrm{unalign}}$ denotes actions in the trace that cannot be aligned with any step, $s^{\mathrm{fail}}$ denotes the step that may be responsible for the failure, and $y \in \{0,1\}$ denotes whether the task~is~successfully completed. These summaries provide compact step-level evidence for diagnosing and updating experience fields in the current~runtime~guidance.

\textbf{Experience Field Diagnosis.}
Given all the extracted action patterns from traces and trace summaries, the miner diagnoses the experience fields, \ie \emph{logical\_actions}, \emph{on\_enter}, and \emph{failure\_patterns}, in the current runtime guidance $\mathcal{G}$. 
Let $\mathcal{F}(\mathcal{G})$ be the set of these fields. For each field $f \in \mathcal{F}(\mathcal{G})$, the miner assigns $d_f =
\mathrm{Diagnose}(f, \{\Sigma(\tau, \mathcal{G}) \mid \tau \in \mathcal{T}\}, \mathcal{P}_{+}, \mathcal{P}_{-})$, where $d_f \in \{\mathrm{FP}, \mathrm{FN}, \mathrm{MC}, \mathrm{OK}\}$, \ly{jointly examining the current field, the trace summaries, and the candidate patterns under the corresponding step context.} Here, $\mathrm{FP}$ means that the field is too restrictive and may incorrectly discourage actions observed in successful traces. $\mathrm{FN}$ means that the field is too weak and fails to capture risky actions observed in failed traces. $\mathrm{MC}$ means missing coverage, where a step lacks sufficient \revision{logical actions in the runtime guidance} for reliable progress tracking. $\mathrm{OK}$ means that no update is needed for the field.

\ly{\textbf{Experience Field Edition.} 
Based on the diagnosis, the miner generates step-level suggestions and warnings, and edits the corresponding experience fields in $\mathcal{G}$. Each field-level edit is represented as an operation $o_f \in \{\mathrm{ADD},\mathrm{UPDATE},\mathrm{REMOVE}\}$. For an $\mathrm{FP}$ diagnosis, it applies $\mathrm{REMOVE}$ or $\mathrm{UPDATE}$ to remove or relax the overly restrictive content. For an $\mathrm{FN}$ diagnosis, it applies $\mathrm{ADD}$ to introduce new action patterns, suggestions, or warnings. It also applies $\mathrm{UPDATE}$ to strengthen the existing content in the corresponding field. For an $\mathrm{MC}$ diagnosis, it applies $\mathrm{ADD}$ to incorporate new action patterns, suggestions, or warnings into the corresponding field. No edit is generated for a field diagnosed as $\mathrm{OK}$. Applying the generated edits to $\mathcal{G}$ produces the populated runtime guidance~$\mathcal{G}^{*}$.}

\ly{\textbf{Guidance Validation.} 
Finally, \tool validates the mined experience against the DSL and the extracted skill specification. It checks that all experience fields are attached to existing steps, each action pattern is well-formed, and each command-matching expression is a valid regular expression. It further checks that runtime guidance does not include duplicated or conflicting entries. If any validation check fails, the errors are returned to the miner for revision.}

\subsection{Step-Aware Runtime Assurance}\label{sec:runtime-assurance}

Given the runtime guidance \revision{$\mathcal{G}^{*}$} for a skill, \tool provides step-aware runtime assurance by wrapping around the agent execution loop through runtime hooks. This online assurance mechanism, which contains \ly{a \textit{procedure checker} and \textit{termination checker}, intervenes when the agent deviates from the skill procedure and provides step-level suggestions and warnings to mitigate errors in individual step execution.}

\textbf{Procedure Checker.}
For each task query, the agent executes the task through an iterative loop of planning, action execution, observation, and re-planning. \tool attaches to this loop through runtime hooks~\cite{claude_code_hooks,codex_hooks}. Before each action planned by the agent is executed, \ly{\tool intercepts the action through a pre-action hook and checks it against \revision{$\mathcal{G}^{*}$}.}

To track execution progress, \tool constructs a finite-state machine (FSM) from the steps and dependency relations in \revision{$\mathcal{G}^{*}$}. Each skill step $s_i \in \mathcal{G}^{*}$ corresponds to an FSM state, and its \emph{depends\_on} field determines when the state can be activated. Let $\mathrm{LA}(s_i)$ denote the set of logical actions associated with $s_i$, and let $\mathrm{M}(s_i) \subseteq \mathrm{LA}(s_i)$ denote the logical actions that have been matched by the observed execution trace. Initially, $\mathrm{M}(s_i)=\emptyset$. A logical action is considered matched when the action planned by the agent conforms to any of its alternative action patterns. $s_i$ is marked as completed only after all of its logical actions have been matched, \ie $\mathrm{M}(s_i)=\mathrm{LA}(s_i)$. Completing a step may subsequently activate other steps whose dependencies~are~satisfied.

\ly{Based on the current FSM state and runtime guidance, \tool handles each action planned by the agent as follows. When the execution enters an activated step for the first time, \tool injects the corresponding \emph{on\_enter} suggestions and warnings into the agent context, \revision{providing step-specific guidance for more reliable step-level execution}. If the action matches \revision{a logical action} of an activated step, \tool allows the action for execution. If it matches a \revision{logical action} belonging to a step whose dependencies have not yet been satisfied, \tool temporarily denies the action, and provides a hint to the LLM agent for re-planning, \revision{avoiding deviations from the expected skill procedure}.}

\ly{\revision{Besides}, the procedure checker also uses the \emph{failure\_patterns} stored in $\mathcal{G}^{*}$ to identify actions associated with historical failures. For each failure-associated action pattern in $\mathcal{G}^{*}$, \tool records whether it has already been triggered during the current execution. When an action matches \revision{a failure-associated action pattern} for the first time, \tool temporarily denies the action and returns a hint containing the corresponding failure reason, thereby giving the agent an opportunity to reconsider its plan. If the agent subsequently generates the same action again, \tool allows it~to~proceed. This design treats execution experience as advisory evidence \revision{for executing individual steps correctly}, and avoids permanently blocking actions that may be valid under the current task.}

\ly{An action that matches neither a logical action nor a failure pattern is also allowed, because it may represent auxiliary exploration or a valid implementation not yet covered by the runtime guidance. However, such an action does not advance the FSM. Every allowed action is executed normally, and its observation is returned to the agent for subsequent planning. Through this process, the procedure checker monitors execution progress, delivers step-level guidance at the corresponding execution context, and intervenes only when it observes a procedural deviation or a failure-associated action pattern.}

\textbf{Termination Checker.}
When the agent signals task completion, the termination checker checks whether all required steps specified in the \emph{termination} field of $\mathcal{G}^{*}$ have been completed. If any required step is missing, it denies the termination request and provides a hint asking the LLM agent to re-plan. Otherwise, it accepts the final output and terminates the task. 

\ly{Finally, \tool follows a fail-open principle~\cite{owaspImproperErrorHandling} to avoid disrupting the underlying agent runtime. If an exception occurs within either checker, it allows the action or termination request by default, and records the error for subsequent diagnosis. \revision{All the agent execution traces of task queries under the current runtime guidance are collected as new inputs for execution experience mining and subsequent runtime-guidance refinement, supporting the self-evolution of \tool.}}

\section{Evaluation}\label{sec:evaluation}
We implement a prototype of \tool with \todo{5,072}~lines of Python code. To evaluate the effectiveness and efficiency of \tool, we design the following research~questions.
\begin{itemize}[leftmargin=*]
    \item \ly{\textbf{RQ1 Effectiveness Evaluation.} How effective is \tool in improving the \revision{reliability} of skill execution?}
    \item \ly{\textbf{RQ2 Efficiency Evaluation.} What runtime overhead does \tool introduce during the LLM agent execution?}
    \item \ly{\textbf{RQ3 Self-Evolving Analysis.} Can \tool evolve with newly collected execution traces?}
    \item \ly{\textbf{RQ4 Ablation Study.} How do execution experience and the guidance delivery strategy contribute to the effectiveness?}
    \item \ly{\textbf{RQ5 Generalization Evaluation.} Can runtime guidance constructed with one backbone model improve skill execution with another backbone model under the same agent?}
\end{itemize}
\subsection{Evaluation Setup}
\textbf{Agent and Model Selection.} We evaluate \tool on two widely used LLM agents, \ie Claude Code~\cite{claudecode} and Codex~\cite{codex}. Both agents are designed for multi-step task execution, and can interact with external tools during execution, supporting the incorporation of external skill documents into the execution context. More importantly, both agents provide hook mechanisms that expose key lifecycle points of the agent execution loop, \ly{allowing us to inspect actions planned by the agent and deliver runtime guidance through their extension interface without modifying the underlying agent implementation.} For each agent, we consider two backbone models. Specifically, we use Claude Code with Claude-Haiku-4.5~\cite{claudehaiku45} and Claude-Opus-4.6~\cite{claudeopus46}, and Codex with GPT-5.2~\cite{gpt52} and GPT-5.4~\cite{gpt54}. For fairness, all the agent-model configurations use the same tool execution environment and maximum task execution time \ly{(\ie 30 minutes)}.

\textbf{Dataset Preparation.}
We build our evaluation dataset based on SkillsBench~\cite{li2026skillsbench}, \ly{which contains 87 tasks associated with one user query, a skill document, a task execution environment and an outcome-based deterministic verifier.} SkillsBench is suitable for our evaluation because its skill documents provide procedural knowledge with high quality, \revision{while its deterministic verifiers reproducibly determine whether each task execution succeeds or fails, thereby labeling the resulting trace as successful or failed for execution experience mining}.

Since our study focuses on runtime assurance rather than teaching an LLM agent a completely new skill, we evaluate skills for which the target agent has already demonstrated basic task-solving capability but may still execute the skill unreliably. \ly{We therefore filter the original SkillsBench skills according to the execution capability of the evaluated agent-model configurations. Specifically, for each of the four agent-model configurations, we execute the original query of every SkillsBench task five times with its corresponding curated skill. A task is regarded as executable under a configuration if at least one of the five executions passes its deterministic verifier. Tasks that fail in all five runs are excluded for that configuration, because such consistent failures are more likely to indicate insufficient basic task-solving capability than unreliable skill execution. We then take the intersection of the executable skill sets obtained from the four configurations. This intersection ensures that every retained skill can be executed successfully by all evaluated configurations at least once, allowing subsequent failures to be more reasonably attributed to execution instability. After this filtering process, 15 skills remain from the original SkillsBench. These commonly executable skills constitute the subjects of our evaluation, and their details are shown in Table~\ref{tab:benchmark_overview}.}

\begin{table*}[!t]
  \centering
  \caption{Overview of the Selected Skills from SkillsBench}
  \label{tab:benchmark_overview}
  \begin{adjustbox}{width=0.8\textwidth}
  \begin{tabular}{ll}
    \toprule
    \textbf{Skill Name} &  \textbf{Description} \\
    \midrule
    macroeconomic-timeseries-detrending & Detrend macroeconomic time series and analyze business-cycle components. \\
    excitation-signal-design & Design excitation signals for system identification and control tasks. \\
    fjsp-repair-with-downtime-and-policy & Repair infeasible flexible job-shop schedules under downtime and policy constraints. \\
    powerlifting-coef-calc & Compute powerlifting coefficients and normalized scores. \\
    threejs & Parse Three.js scene graphs and export articulated mesh assets. \\
    weighted-gdp-calc & Calculate weighted GDP indicators from regional or sector-level data. \\
    pddl-skills & Load, solve, validate, and save plans for PDDL planning tasks. \\
    protein-expression-analysis & Analyze protein-related data for biochemical or structural insights. \\
    glm-calibration & Calibrate GLM parameters for water temperature simulation. \\
    pcap-analysis & Analyze PCAP files and compute network statistics. \\
    spring-boot-migration & Migrate Spring Boot 2.x applications to Spring Boot 3.x. \\
    mesh-analysis & Analyze STL meshes to compute geometry and filter scan noise. \\
    jackson-security & Analyze Jackson deserialization security risks and attack patterns. \\
    geospatial-analysis & Analyze geospatial data with proper projections and spatial operations. \\
    d3-visualization & Build deterministic D3.js visualizations from local data. \\
    \bottomrule
  \end{tabular}
  \end{adjustbox}
\end{table*}

Since each retained SkillsBench skill originally provides only one task query, we expand the query set for each selected skill through a two-stage augmentation process inspired by~\cite{li2026no}. \ly{The objective is to increase both task-level and expression-level diversity while preserving deterministic and verifiable task oracles for evaluation.} In the first stage, we perform \textit{task-level semantic expansion}. \ly{For each skill, we construct 8 task instances, including the original task and 7 manually created variants.} These variants preserve the intended skill usage but change concrete query semantics, such as input parameters, target objects, and task context. We manually adapt the corresponding deterministic verifier for each newly constructed instance. In the second stage, we perform \textit{expression-level paraphrase mutation}. \ly{For each task instance, we construct ten semantically equivalent query expressions, including its original expression and nine paraphrases generated using GPT-5.4~\cite{gpt54}. The paraphrases modify only surface-level expressions, such as wording and sentence structure, without changing the task requirements or expected outcome. Therefore, all expression variants derived from the same task instance share the verifiers.}

In total, each skill contains \todo{80} queries, resulting in \todo{1200} queries across 15 \ly{selected} skills. For each skill, \ly{we divide the queries into an evolution set and a held-out test set,~denoted by $\mathcal{Q}_{evol}$ and $\mathcal{Q}_{test}$, respectively. We perform this split at the task-instance level. Five complete task instances, containing 50 queries in total, are assigned to $\mathcal{Q}_{evol}$, while the remaining three task instances, containing 30 queries, are assigned to $\mathcal{Q}_{test}$. Consequently, paraphrases derived from the same task instance never appear in both sets. The 50 queries in $\mathcal{Q}_{evol}$ are used to simulate a stream of new task queries arriving after deployment. \revision{As new queries arrive, their successful and failed execution traces are progressively collected and used to mine execution experience and update the runtime guidance.} The 30 queries in $\mathcal{Q}_{test}$ remain unseen~throughout evolution and are used only for effectiveness evaluation.}

\textbf{RQ Setup.} For \textbf{RQ1}, \ly{we evaluate whether \tool improves both task success across queries and execution stability under repeated runs. For each skill and agent-model configuration, we evolve the runtime guidance for 10 iterations using the 50 queries in $\mathcal{Q}_{evol}$. In each iteration, we randomly select five previously unused queries, execute them under the current guidance, collect the resulting traces, and update the guidance accordingly. After all 50 queries have been processed, we evaluate the final guidance on the 30 queries in $\mathcal{Q}_{test}$.}

\revision{We evaluate each evolved guidance five times on $\mathcal{Q}_{test}$ and report the mean and standard deviation (SD) of the resulting task success rates for each skill and agent-model configuration, where a higher mean and lower SD indicate more reliable execution of LLM agents with the corresponding skill}.

For \textbf{RQ2}, \ly{we evaluate the runtime overhead introduced by \tool on $\mathcal{Q}_{test}$. We compare the base agent-model configuration and \tool in terms of the average number of inference turns and token costs. We further report the computation time of \tool, and its proportion of the total execution time for each agent-model~configuration.}

\ly{For \textbf{RQ3}, we analyze the self-evolving capability of \tool by evaluating it after each of the 10 evolution \revision{iterations}. At the end of each \revision{iteration}, we evaluate \tool on $\mathcal{Q}_{test}$ and compute the task success rate of each skill. To reduce the effects of query ordering, we repeat the entire evolution process five times. We aggregate the results over five independent evolution processes, and report how the average task success rate changes as more traces are included.}

\ly{For \textbf{RQ4}, we conduct ablation studies across all four agent-model configurations to examine the contributions of execution experience and step-level guidance delivery. After the \revision{10-iteration} evolution process, we construct three variants. \emph{w/o On-Enter} removes the \emph{on\_enter} field, including step-level suggestions and warnings; \emph{w/o Failure-Patterns} removes the \emph{failure\_patterns} field; and \emph{System-Prompt Delivery} provides the complete runtime guidance in the system prompt at the beginning of execution instead of delivering the corresponding guidance at runtime. We evaluate all variants on $\mathcal{Q}_{test}$ and compare their task success rates with \tool.}

\ly{For \textbf{RQ5}, we evaluate the cross-model generalization of the evolved runtime guidance within the same agent. We consider bidirectional transfer between Claude-Haiku-4.5 and Claude-Opus-4.6, and between GPT-5.2 and GPT-5.4. We directly apply the guidance evolved with the source backbone model to the target model without further updates, and compare its task success rate on $\mathcal{Q}_{test}$ with those of the target-model baseline and the guidance evolved natively with the target model.}

\textbf{Environment.} We conduct all the experiments on Ubuntu 20.04.4 LTS servers with 4 NVIDIA GeForce RTX 3090 GPUs, Intel(R) Xeon(R) Silver 4310 @ 2.10GHz and 128GB memory.

\subsection{Effectiveness Evaluation (RQ1)}\label{sec:effectiveness-evaluation}

\begin{table*}[t]
\centering
\caption{Results of Effectiveness Evaluation}
\vspace{-5pt}
\label{tab:task_success_rate}
\begin{adjustbox}{width=\textwidth}
\begin{tabular}{l*{8}{c}}
\toprule
\multirow{2}{*}{\textbf{Skill}}
& \multicolumn{2}{c}{\textbf{Claude Code + Haiku-4.5}}
& \multicolumn{2}{c}{\textbf{Claude Code + Opus-4.6}}
& \multicolumn{2}{c}{\textbf{Codex + GPT-5.2}}
& \multicolumn{2}{c}{\textbf{Codex + GPT-5.4}} \\
\cmidrule(lr){2-3}
\cmidrule(lr){4-5}
\cmidrule(lr){6-7}
\cmidrule(lr){8-9}
& \textbf{Base Agent} & \textbf{\tool}
& \textbf{Base Agent} & \textbf{\tool}
& \textbf{Base Agent} & \textbf{\tool}
& \textbf{Base Agent} & \textbf{\tool} \\
\midrule
macroeconomic-timeseries-detrending & $0.613\pm0.018$ & $\mathbf{0.893}\pm\mathbf{0.015}$ & $0.800\pm0.033$ & $\mathbf{0.933}\pm\mathbf{0.024}$ & $0.827\pm0.028$ & $\mathbf{0.893}\pm\mathbf{0.015}$ & $0.833\pm0.033$ & $\mathbf{0.947}\pm\mathbf{0.018}$ \\
excitation-signal-design & $0.120\pm\mathbf{0.030}$ & $\mathbf{0.373}\pm0.037$ & $0.187\pm0.030$ & $\mathbf{0.487}\pm\mathbf{0.018}$ & $0.213\pm0.045$ & $\mathbf{0.387}\pm\mathbf{0.018}$ & $0.240\pm0.028$ & $\mathbf{0.513}\pm\mathbf{0.018}$ \\
fjsp-repair-with-downtime-and-policy & $0.707\pm\mathbf{0.028}$ & $\mathbf{0.773}\pm0.037$ & $0.733\pm0.033$ & $\mathbf{0.860}\pm\mathbf{0.015}$ & $0.540\pm0.028$ & $\mathbf{0.707}\pm\mathbf{0.015}$ & $0.607\pm0.028$ & $\mathbf{0.793}\pm\mathbf{0.015}$ \\
powerlifting-coef-calc & $0.653\pm0.051$ & $\mathbf{0.867}\pm\mathbf{0.024}$ & $0.840\pm0.028$ & $\mathbf{0.933}\pm\mathbf{0.024}$ & $0.527\pm0.028$ & $\mathbf{0.760}\pm0.028$ & $0.673\pm0.037$ & $\mathbf{0.893}\pm\mathbf{0.015}$ \\
threejs & $0.420\pm0.030$ & $\mathbf{0.560}\pm\mathbf{0.015}$ & $0.567\pm\mathbf{0.033}$ & $\mathbf{0.673}\pm0.037$ & $0.473\pm0.037$ & $\mathbf{0.607}\pm\mathbf{0.015}$ & $0.560\pm0.037$ & $\mathbf{0.700}\pm\mathbf{0.033}$ \\
weighted-gdp-calc & $0.267\pm0.033$ & $\mathbf{0.573}\pm\mathbf{0.015}$ & $0.313\pm0.030$ & $\mathbf{0.600}\pm\mathbf{0.024}$ & $0.280\pm0.051$ & $\mathbf{0.487}\pm\mathbf{0.018}$ & $0.247\pm\mathbf{0.030}$ & $\mathbf{0.540}\pm0.043$ \\
pddl-skills & $0.213\pm0.030$ & $\mathbf{0.373}\pm\mathbf{0.015}$ & $0.360\pm0.037$ & $\mathbf{0.607}\pm\mathbf{0.015}$ & $0.407\pm0.028$ & $\mathbf{0.633}\pm\mathbf{0.024}$ & $0.513\pm\mathbf{0.030}$ & $\mathbf{0.667}\pm0.033$ \\
protein-expression-analysis & $0.840\pm0.043$ & $\mathbf{0.920}\pm\mathbf{0.018}$ & $0.827\pm0.028$ & $\mathbf{0.947}\pm\mathbf{0.018}$ & $0.693\pm\mathbf{0.043}$ & $\mathbf{0.853}\pm0.045$ & $0.880\pm0.018$ & $\mathbf{0.920}\pm0.018$ \\
glm-calibration & $0.887\pm0.038$ & $\mathbf{1.000}\pm\mathbf{0.000}$ & $0.913\pm0.030$ & $\mathbf{1.000}\pm\mathbf{0.000}$ & $0.880\pm0.018$ & $\mathbf{1.000}\pm\mathbf{0.000}$ & $0.927\pm0.028$ & $\mathbf{1.000}\pm\mathbf{0.000}$ \\
pcap-analysis & $0.893\pm0.043$ & $\mathbf{0.967}\pm\mathbf{0.024}$ & $0.940\pm0.015$ & $\mathbf{1.000}\pm\mathbf{0.000}$ & $0.980\pm0.018$ & $\mathbf{1.000}\pm\mathbf{0.000}$ & $0.973\pm0.015$ & $\mathbf{1.000}\pm\mathbf{0.000}$ \\
spring-boot-migration & $0.920\pm0.018$ & $\mathbf{0.973}\pm\mathbf{0.015}$ & $0.927\pm0.028$ & $\mathbf{1.000}\pm\mathbf{0.000}$ & $0.973\pm0.015$ & $\mathbf{1.000}\pm\mathbf{0.000}$ & $0.980\pm0.018$ & $\mathbf{1.000}\pm\mathbf{0.000}$ \\
mesh-analysis & $0.847\pm0.038$ & $\mathbf{0.953}\pm\mathbf{0.018}$ & $0.913\pm0.038$ & $\mathbf{1.000}\pm\mathbf{0.000}$ & $0.840\pm0.037$ & $\mathbf{1.000}\pm\mathbf{0.000}$ & $0.887\pm0.030$ & $\mathbf{0.967}\pm\mathbf{0.024}$ \\
jackson-security & $0.433\pm0.024$ & $\mathbf{0.613}\pm\mathbf{0.018}$ & $0.600\pm\mathbf{0.033}$ & $\mathbf{0.700}\pm0.053$ & $0.467\pm0.024$ & $\mathbf{0.640}\pm\mathbf{0.015}$ & $0.653\pm0.018$ & $\mathbf{0.707}\pm\mathbf{0.015}$ \\
geospatial-analysis & $0.640\pm0.037$ & $\mathbf{0.900}\pm\mathbf{0.024}$ & $0.767\pm0.024$ & $\mathbf{0.947}\pm\mathbf{0.018}$ & $0.793\pm\mathbf{0.028}$ & $\mathbf{0.900}\pm0.033$ & $0.800\pm0.024$ & $\mathbf{0.947}\pm\mathbf{0.018}$ \\
d3-visualization & $0.127\pm0.037$ & $\mathbf{0.380}\pm\mathbf{0.018}$ & $0.233\pm0.033$ & $\mathbf{0.473}\pm\mathbf{0.028}$ & $0.160\pm0.028$ & $\mathbf{0.420}\pm\mathbf{0.018}$ & $0.247\pm0.051$ & $\mathbf{0.487}\pm\mathbf{0.018}$ \\
\midrule
\textbf{Average} & 0.572 & \textbf{0.741} & 0.661 & \textbf{0.811} & 0.604 & \textbf{0.752} & 0.668 & \textbf{0.805} \\
\bottomrule
\end{tabular}%
\end{adjustbox}
\end{table*}

\textbf{Overall Effectiveness.}
Table~\ref{tab:task_success_rate} shows the mean task~success rate and standard deviation over five repeated evaluations for the base agents and \tool across 15 skills and four agent-model configurations. The higher mean and lower standard deviation are highlighted in bold. Overall, \tool consistently improves task success \revision{rates} across all evaluated settings. Averaged across the 15 skills, \tool achieves relative improvements of \todo{29.5\%} and \todo{22.7\%} for Claude Code with Claude-Haiku-4.5 and Claude-Opus-4.6, respectively, and \todo{24.5\%} and \todo{20.5\%} for Codex with GPT-5.2 and GPT-5.4, respectively. Across all four configurations, \tool increases the average task success rate from \todo{62.6\%} to \todo{77.7\%}, corresponding to an overall relative improvement of \todo{24.1\%}.

The improvement is consistent across all agent-model~configurations and skills. The gains are particularly pronounced on skills with low baseline performance, reaching \todo{210.8\%} for \textit{excitation-signal-design}, \todo{131.6\%} for \textit{weighted-gdp-calc}, and \todo{199.2\%} for \textit{d3-visualization}, with substantial gains also observed for \textit{pddl-skills} and \textit{geospatial-analysis}. For skills with strong baseline performance, the relative gains are smaller, yet \tool still reduces residual failures, achieving perfect task success rates for \textit{glm-calibration} across all four configurations and for several configurations of \textit{pcap-analysis}, \textit{spring-boot-migration}, and \textit{mesh-analysis}. Overall, \tool improves challenging skill executions while mitigating occasional failures on skills that already succeed in most runs.

In addition to improving task success rates, \tool generally produces more stable results across repeated evaluations. The average standard deviation decreases by \todo{41.2\%}, \todo{39.5\%}, \todo{46.5\%}, and \todo{36.9\%} under Claude-Haiku-4.5, Claude-Opus-4.6, GPT-5.2, and GPT-5.4, respectively. Across all 60 skill-agent-model configurations, \tool yields a lower standard deviation in 50 pairs and the same standard deviation in two pairs. Overall, it reduces the average standard deviation by \todo{41.1\%}, indicating that these effectiveness improvements are accompanied by more consistent execution across repeated~runs.

\begin{figure}[!t]
  \centering \includegraphics[width=0.9\linewidth]{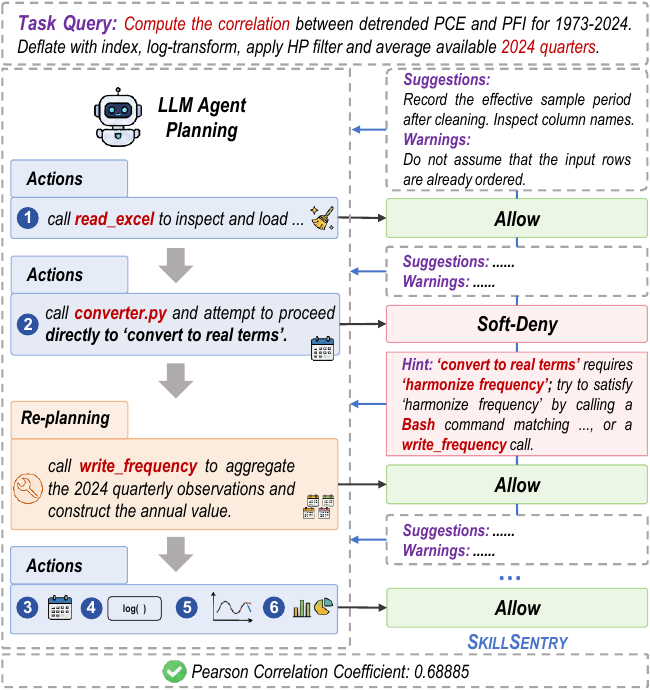} 
  \caption{An Example of Runtime Assurance by \tool} 
  \label{fig:rq1-example} 
\end{figure}

\textbf{Case Study.}
Fig.~\ref{fig:rq1-example} illustrates how \tool assures Claude Code with Claude-Haiku-4.5 when executing the \textit{macroeconomic-timeseries-detrending} skill. Upon entering each skill step, \tool provides the corresponding suggestions and warnings from the runtime guidance. After loading and inspecting the data, the agent attempts to proceed directly to \textit{Convert to Real Terms} without completing the required preceding step, \ie \textit{Harmonize Frequency}. \tool detects the deviation from the skill procedure and temporarily denies the action. It returns a targeted hint identifying the missing step. Guided by this intervention, the agent re-plans and invokes \texttt{write\_frequency} to aggregate the quarterly observations and construct the annual value, thereby completing the skipped step. Then, \tool allows execution to continue, and verifies that all required steps are completed before accepting the final output. This example shows how \tool combines step-level guidance and runtime intervention to improve the reliability of agent execution.

\textit{\textbf{Summary.}} \tool consistently improves the reliability of skill execution across all evaluated agents, backbone models, and skills. Overall, \tool improves the average task success rate by \todo{24.1\%} and reduces the average standard deviation across repeated evaluations by \todo{41.1\%}.

\subsection{Efficiency Evaluation (RQ2)}\label{sec:efficiency-evaluation}
\begin{table}[t]
    \centering
    \caption{Results of Runtime Cost}
    \label{tab:runtime_cost}
    \begin{adjustbox}{width=\linewidth}
    \begin{tabular}{lcccc}
        \toprule
        \multirow{2}{*}{\textbf{Agent-Model}}
        & \multicolumn{2}{c}{\textbf{Inference Turn}}
        & \multicolumn{2}{c}{\textbf{Token Cost}}
        \\
        \cmidrule(lr){2-3}
        \cmidrule(lr){4-5}
        & \textbf{Base Agent}
        & \textbf{\tool}
        & \textbf{Base Agent}
        & \textbf{\tool}
        \\
        \midrule
        Claude Code + Haiku-4.5
        & 20.2
        & 22.2
        & 796K
        & 886K
        \\

        Claude Code + Opus-4.6
        & 24.3
        & 25.3
        & 1.30M
        & 1.38M
        \\

        Codex + GPT-5.2
        & 23.7
        & 25.6
        & 673K
        & 747K
        \\

        Codex + GPT-5.4
        & 27.8
        & 30.4
        & 757K
        & 804K
        \\
        \bottomrule
    \end{tabular}
    \end{adjustbox}
\end{table}

Table~\ref{tab:runtime_cost} reports the runtime costs of the base agents and \tool. Averaged across the four agent-model configurations, \tool increases the number of inference turns by \todo{7.8\%} and the token cost by \todo{8.7\%}. These additional costs mainly arise from delivering step-level guidance and prompting the agent to re-plan when deviations from the skill procedure are detected. Therefore, these additional inference costs represent a reasonable trade-off for correcting unreliable execution and achieving the effectiveness improvements reported in \textbf{RQ1}.

\begin{figure}[!t]
  \centering \includegraphics[width=0.9\linewidth]{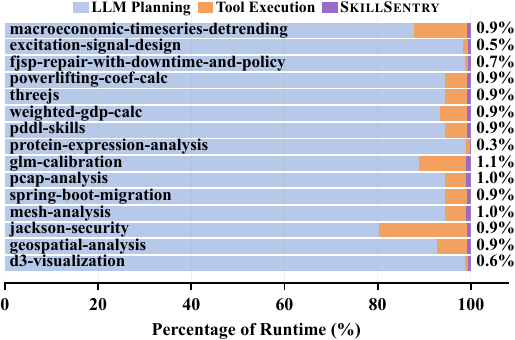} 
  \caption{Results of the Runtime Efficiency} 
  \label{fig:rq2-efficiency} 
\end{figure}

Fig.~\ref{fig:rq2-efficiency} presents the runtime breakdown for Claude Code with Claude-Haiku-4.5. The computation performed by \tool accounts for approximately \todo{0.8\%} of the total execution time. This result indicates that \tool introduces negligible computational overhead compared with LLM inference and tool execution. The runtime breakdowns for the other three agent-model configurations are similar~and~are~available~at~\cite{website}.

\textit{\textbf{Summary.}}
\tool introduces \todo{7.8\%} more inference turns and \todo{8.7\%} more token costs, while its own computation accounts for only \todo{0.8\%} of the total execution~time. 
\subsection{Self-Evolving Analysis (RQ3)}\label{sec:self-evolving-analysis}

\begin{figure}[t] 
  \centering \includegraphics[width=0.9\linewidth]{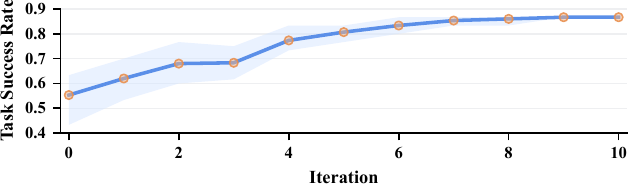} 
  \caption{Results of Self-Evolving Analysis} 
  \label{fig:rq3-self-evolving} 
\end{figure}

Fig.~\ref{fig:rq3-self-evolving} illustrates the self-evolving process of \tool for \textit{macroeconomic-timeseries-detrending} under Claude Code with Claude-Haiku-4.5. \revision{The shaded area denotes the standard deviation~(SD) across five repeated evolution runs.} As more traces are incorporated, the average task success rate improves by about \todo{58.2\%} after 10 \revision{iterations}. The largest gains occur in the early stages, as newly collected traces enrich action patterns, suggestions, and warnings. The improvement becomes relatively small during later stages, indicating that the runtime guidance tends to stabilize on the current test set. Meanwhile, the \revision{decreasing SD} across repeated runs suggests increasingly stable performance of \tool as more traces are incorporated. Similar trends are observed for the other skills and agent-model configurations, whose~results~are~available~at~\cite{website}.

\textit{\textbf{Summary.}}
\tool progressively improves task success rate as newly collected traces are incorporated, demonstrating that its runtime guidance can self-evolve and converge toward more reliable skill execution.
\subsection{Ablation Study (RQ4)}\label{sec:ablation-study}

\begin{table*}[t]
\centering
\caption{Results of Ablation Study}
\label{tab:ablation}
\begin{adjustbox}{width=0.8\linewidth}
\begin{tabular}{lccccc}
\toprule
\textbf{Agent--Model}
& \textbf{Base Agent}
& \textbf{w/o On-Enter}
& \textbf{w/o Failure Patterns}
& \textbf{System-Prompt}
& \textbf{\tool} \\
\midrule
Claude Code + Haiku-4.5 & $0.572\pm0.033$ & $0.600\pm0.024$ & $0.636\pm0.021$ & $0.693\pm0.030$ & $\mathbf{0.741}\pm\mathbf{0.019}$ \\
Claude Code + Opus-4.6 & $0.661\pm0.030$ & $0.692\pm0.021$ & $0.719\pm0.019$ & $0.768\pm0.023$ & $\mathbf{0.811}\pm\mathbf{0.018}$ \\
Codex + GPT-5.2 & $0.604\pm0.030$ & $0.627\pm0.022$ & $0.661\pm0.017$ & $0.710\pm0.022$ & $\mathbf{0.752}\pm\mathbf{0.016}$ \\
Codex + GPT-5.4 & $0.668\pm0.028$ & $0.693\pm0.021$ & $0.721\pm0.017$ & $0.768\pm0.022$ & $\mathbf{0.805}\pm\mathbf{0.018}$ \\
\bottomrule
\end{tabular}
\end{adjustbox}
\end{table*}

Table~\ref{tab:ablation} reports the ablation results across the four agent-model configurations. Overall, the complete \tool consistently achieves the highest average task success rate. \tool outperforms \emph{w/o On-Enter} and \emph{w/o Failure Patterns} by \todo{19.0\%} and \todo{13.6\%}, respectively. This result confirms that both forms of execution experience improve skill reliability. Removing \emph{on\_enter} causes a larger degradation, highlighting the importance of step-level suggestions and warnings, while failure-associated action patterns provide complementary guidance by prompting agents to reconsider previously unsuccessful~actions, helping the execution of~individual~steps.

\tool also outperforms \emph{System-Prompt Delivery}~by \todo{5.8\%} on average. Although both use the same guidance, step-level delivery is more effective because it presents relevant information in the corresponding execution context, reducing the chance that suggestions or warnings are overlooked in a long initial prompt. Moreover, the complete \tool also achieves the lowest average standard deviation, indicating improvements in both effectiveness and stability.

\textit{\textbf{Summary.}}
Step-level suggestions, warnings, and failure-associated action patterns all contribute to the effectiveness of \tool. Delivering guidance at the step level further improves the average task success rate by \todo{5.8\%}.
\subsection{Generalization Evaluation (RQ5)}\label{sec:generalization-evaluation}

\begin{figure}[t] 
    \centering 
    \includegraphics[width=0.9\linewidth]{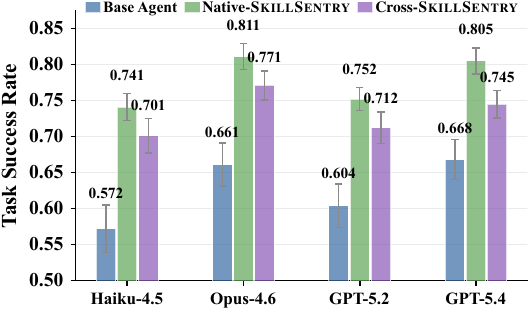} 
    \caption{Results of the Generalization Evaluation} 
    \label{fig:rq5-generalization} 
\end{figure}

Fig.~\ref{fig:rq5-generalization} reports the cross-model generalization results. Native-\tool uses runtime guidance evolved with the target backbone model, whereas Cross-\tool directly transfers guidance evolved with another model under the same agent. Cross-\tool outperforms the base agent, improving the task success rate by \todo{22.6\%}, \todo{16.6\%}, \todo{17.9\%}, and \todo{11.5\%} for Claude-Haiku-4.5, Claude-Opus-4.6, GPT-5.2, and GPT-5.4, respectively, with an average relative improvement~of~\todo{17.2\%}.

Although Cross-\tool remains below Native-\tool, it retains \todo{94.2\%} of the native task success rate on average. This result suggests that much of the mined execution experience, including action patterns, step-level suggestions, and warnings, transfers across backbone models within the same agent, while the remaining gap reflects model-specific action patterns and failure behaviors. In both model pairs, transferring guidance from the stronger backbone to the weaker one yields larger improvements than the reverse~transfer.

\textit{\textbf{Summary.}}
Runtime guidance evolved with one backbone model generalizes to another within the same agent, improving task success rate by \todo{17.2\%} over the base agent while retaining \todo{94.2\%} of the effectiveness of natively evolved~guidance.

\section{Threats to Validity}\label{sec:threats}

First, the selection of skills, agents, and backbone models may limit generalizability. To mitigate this threat, we evaluate the 15 skills across two widely used agents, \ie Claude Code and Codex, each paired with two backbone models.

Second, the original capabilities of the agents and the quality of skill documents pose threats to validity. \tool assumes that the agent correctly selects the required skill and the procedure in the skill document is correct. It focuses on assuring individual skill execution rather than skill selection. When a task requires multiple skills, \tool can support their execution when the skills are invoked sequentially. However, it does not currently provide effective assurance for workflows that repeatedly switch between interdependent steps from different skills. Such cases may require integrating and jointly optimizing the involved skills into a unified one.

Third, execution experience mining relies on a verifier to label successful and failed traces. We use deterministic verifiers provided by the evaluation dataset to ensure reliable trace labels in our experiments. In practice, such verifiers may require additional construction effort. In these cases, \tool can instead incorporate LLM-based evaluators tailored to the target task for agent execution trace labeling.

Finally, the stochasticity of LLM-based execution and randomness of query ordering in experiments pose threats to validity. To mitigate these threats, we evaluate each evolved guidance five times, repeat the evolution process five times, and report both mean and standard deviation. We also conduct paired $t$-test analysis~\cite{box1987guinness} across our comparisons, and the improvements are statistically significant (p$<$0.05).


\section{Related Work}\label{sec:related-work}

\subsection{Runtime Assurance for LLM Agents}
The autonomy of LLM agents brings new safety and~reliability challenges, motivating recent work~\cite{wang2026agentspec,chen2025shieldagent,kumar2026infrastructuresentinel} on runtime guardrails, monitoring, and enforcement. \textsc{AgentSpec}~\cite{wang2026agentspec} allows users to specify runtime constraints for LLM agents with a domain-specific language and enforces these constraints during execution. \textsc{ShieldAgent}~\cite{chen2025shieldagent} extracts verifiable rules from policy documents and uses them to check whether~the trace of agent actions violates safety policies. Kumar~et~al.~\cite{kumar2026infrastructuresentinel} further study policy-enforced~guardrails for~infrastructure~agents. These studies show the importance of adding an external runtime module to constrain agent behaviors without modifying the underlying model. \ly{\tool differs from these studies by targeting skill execution reliability rather than safety enforcement. To the best of our knowledge, we are the first to propose a runtime assurance framework for skill-oriented agent execution. We therefore do not directly compare \tool with these approaches, as prior approaches detect safety violations, whereas \tool enforces skill procedures and leverages execution experience to improve task completion, resulting in different inputs, objectives, and evaluation criteria.}

\subsection{Execution-Based Optimization for LLM Agents}
A growing body of work improves LLM agents by optimizing or augmenting system prompts, agent memories, and skills based on execution feedback. Reflexion~\cite{shinn2023reflexion} stores verbal reflections generated from task feedback and reuses them in~later trials. ExpeL~\cite{zhao2023expel} summarizes successful and failed traces~into reusable natural-language experience in the system prompt. APE~\cite{zhou2023large}~generates instruction candidates with LLMs and selects prompts according to task performance. ProTeGi~\cite{pryzant2023automatic} edits prompts~using textual gradients derived from model errors, while OPRO~\cite{yang2024large} treats prompt optimization as a natural-language optimization process driven by LLM-generated proposals. Recently, Trace2Skill~\cite{ni2026trace2skill} distills prior failure lessons into transferable agent skills, providing a new perspective on~leveraging execution traces for agent improvement. Instead of rewriting a prompt or skill artifact offline, \tool uses execution traces to support runtime assurance. It aligns successful and failed traces with skill steps, mines step-level runtime guidance, and provides such guidance only when the agent reaches the corresponding~step during execution, improving the runtime~reliability of skill execution.


\section{Conclusion}\label{sec:conclusion}
We have proposed and implemented \tool, a skill-oriented runtime assurance framework for improving the reliability of LLM agents. Our evaluation shows that it effectively improves execution reliability with low runtime~overhead.



{\footnotesize
\bibliographystyle{IEEEtranS}
\bibliography{IEEEabrv,src/reference}
}

\end{document}